\pdfoutput=1

\documentclass[11pt]{article}

\usepackage[final]{acl}

\usepackage{times}
\usepackage{latexsym}

\usepackage[T1]{fontenc}

\usepackage[utf8]{inputenc}

\usepackage{microtype}

\usepackage{inconsolata}

\usepackage{graphicx}

\usepackage{amsfonts}
\usepackage{bm}
\usepackage{marvosym}
\usepackage{multirow}
\usepackage{booktabs}
\usepackage{xcolor}

\usepackage{amsmath}
\usepackage{amsfonts}
\usepackage{stfloats}
\usepackage{float}
\usepackage{subfig}
\usepackage[normalem]{ulem}
\useunder{\uline}{\ul}{}
\usepackage{hyperref}

\title{JOLT-SQL: Joint Loss Tuning of Text-to-SQL with Confusion-aware Noisy Schema Sampling}

 \author{
 \textbf{Jinwang Song}\textsuperscript{1}, 
 \textbf{Hongying Zan}\textsuperscript{1}\thanks{Corresponding author.}, 
 \textbf{Kunli Zhang}\textsuperscript{1},
 \textbf{Lingling Mu}\textsuperscript{1},
 \textbf{Yingjie Han}\textsuperscript{1}, \\
 \textbf{Haobo Hua}\textsuperscript{2},
 \textbf{Min Peng}\textsuperscript{3}, 
 \\
$^{1}$Zhengzhou University,
$^{2}$Zhengzhou University of Aeronautics,
$^{3}$Wuhan University, \\
\texttt{jwsong@gs.zzu.edu.cn, iehyzan@zzu.edu.cn}
}

\begin{document}
\maketitle
\begin{abstract}
Text-to-SQL, which maps natural language to SQL queries, has benefited greatly from recent advances in Large Language Models (LLMs). While LLMs offer various paradigms for this task, including prompting and supervised fine-tuning (SFT), SFT approaches still face challenges such as complex multi-stage pipelines and poor robustness to noisy schema information. To address these limitations, we present JOLT-SQL, a streamlined single-stage SFT framework that jointly optimizes schema linking and SQL generation via a unified loss. JOLT-SQL employs discriminative schema linking, enhanced by local bidirectional attention, alongside a confusion-aware noisy schema sampling strategy with selective attention to improve robustness under noisy schema conditions. Experiments on the Spider and BIRD benchmarks demonstrate that JOLT-SQL achieves state-of-the-art execution accuracy among comparable-size open-source models, while significantly improving both training and inference efficiency. Our code is available at \href{https://github.com/Songjw133/Joint-Loss-Tuning-of-Text-to-SQL}{https://github.com/Songjw133/JOLT-SQL}.
\end{abstract}

\section{Introduction}

Text-to-SQL technology, which transforms natural language questions into executable SQL queries \citep{deng2022recent}, aims to break down the technical barriers for users interacting with complex databases, enabling non-professionals to conveniently access desired data. With the widespread adoption of data-driven applications across various domains, the research value and application prospects of Text-to-SQL have become increasingly significant. In recent years, the rise of Large Language Models (LLMs) has injected new vitality into this field, and LLM-based Text-to-SQL methods have made remarkable progress \citep{survey-1,survey-2}.

Currently, LLM-based Text-to-SQL methods can be broadly categorized into two main paradigms: Prompting approaches, including methods like In-Context Learning (ICL) and Chain-of-Thought (CoT) \citep{cot,icl}; and Supervised Fine-Tuning (SFT). Prompting methods have garnered considerable attention due to their impressive performance in zero-shot or few-shot scenarios. However, they often rely on powerful, proprietary commercial models, and their effectiveness is typically difficult to replicate with smaller-parameter open-source models.

In contrast, SFT offers a more controllable path to enhance open-source models. Task-specific SFT significantly improves SQL generation capabilities and provides practical advantages such as lower deployment costs, offline operation, and support for local data processing.

Schema linking is a crucial component in mainstream Text-to-SQL methods \citep{yang2024sql}. Its purpose is to select the necessary database tables/columns and filter out database schema elements irrelevant to the user's question, thereby reducing interference and improving SQL generation performance. CHESS \citep{chess} employs a dedicated agent to prune irrelevant schema elements. E-SQL \citep{e-sql} optimizes the schema linking phase through question enrichment and candidate predicate expansion. DIN-SQL \citep{din-sql} enhances the model's schema linking capabilities using decomposed ICL.

In existing approaches, schema linking is often treated as a separate step from SQL generation. In SFT-based methods, schema linking and SQL generation tasks are typically trained separately \citep{dts-sql}, leading to increased task costs. In contrast, ROUTE \citep{route} utilizes multi-task learning to incorporate schema linking, SQL generation, and other sub-tasks into a single SFT process. However, this approach also means that the total volume of training data increases linearly with the number of tasks, introducing additional training overhead.

In this work, we propose a novel method: \textbf{JO}int \textbf{L}oss \textbf{T}uning of Text-to-SQL with Confusion-aware Noisy Schema Sampling (JOLT-SQL). Our objective is to address two primary issues: (1) Conventional pipeline fine-tuning strategies or multi-task learning approaches often require fine-tuning multiple distinct models or incur increased training time due to the expanded multi-task dataset, leading to higher temporal and hardware costs. (2) During actual inference, the schema information provided by the schema linking module is often imperfect, potentially containing irrelevant items or omitting necessary ones. Existing SQL generation models, when fine-tuned solely on ground truth schema or with random sampled noisy schema, struggle to adapt to such imperfect inputs during inference, thereby affecting SQL generation accuracy.

JOLT-SQL addresses these challenges through an innovative joint loss tuning strategy. Specifically, by adjusting attention masks, JOLT-SQL stands as the first LLM SFT method to fuse the optimization objectives of schema linking and SQL generation tasks within a single backward pass. Concurrently, the method dynamically samples noisy schema based on probability during training, enhancing its ability to handle imperfect schema linking results during inference.

Our contributions are as follows:

\(\bullet\) We propose a discriminative schema linking fine-tuning method that incorporates local bidirectional attention. Experiments demonstrate that this method achieves state-of-the-art performance in schema linking on the Spider and BIRD datasets, along with faster inference speeds.

\(\bullet\) Unlike typical pipeline fine-tuning strategies, we train schema linking and SQL generation using joint loss, thereby avoiding the additional overhead associated with multi-stage fine-tuning.

\(\bullet\) We further introduce Confusion-aware Noisy Schema Sampling, enhancing model robustness against redundant schema linking results. This technique strategically guides the SQL generation task's attention by incorporating noisy schema items selected based on the model's points of predictive confusion. Experimental results show that JOLT-SQL, using the Qwen2.5-Coder-14B model, achieves 88.4\%/88.9\% execution accuracy on the Spider Dev/Test set and 64.9\% on the BIRD Dev set.

\section{Related Work}
Large Language Models (LLMs) have shown significant promise in Text-to-SQL, with research largely following two paths: Prompting and Supervised Fine-Tuning (SFT).
\subsection{Prompting for Text-to-SQL}
These methods guide LLMs with specific instructions. DIN-SQL \citep{din-sql} used decomposed in-context learning and a multi-stage pipeline. DAIL-SQL \citep{dail-sql} focused on example selection for complex scenarios. Other works refined prompting and schema understanding: C3 \citep{c3} improved schema representation and used a multi-module pipeline, while MAC-SQL \citep{mac-sql} employed multi-agent collaboration. To handle increased complexity, CHESS \citep{chess} utilized agent-based schema pruning and multi-turn dialogue; E-SQL \citep{e-sql} optimized schema linking via question enrichment and candidate predicate expansion; and RSL-SQL \citep{rsl-sql} applied bi-directional pruning methods to improve schema linking recall. The effectiveness of these approaches remains highly dependent on prompt design and the LLM's inherent understanding capabilities.

\subsection{Supervised Fine-tuning based Text-to-SQL}
SFT aims to more deeply integrate task-specific knowledge into open-source LLMs. While early methods involved direct language model fine-tuning, later strategies became more specialized. DTS-SQL and DB-Explore \citep{dts-sql,db-explore} treated schema linking as a separate fine-tuning stage, whereas ROUTE \citep{route} used multi-task learning for schema linking and SQL generation among other sub-objectives. ExSL \citep{exsl} proposed an efficient extractive schema linking method for decoder-only LLMs. Other approaches include SENSE \citep{sense}, which introduced reinforcement learning, and SQL-PaLM \citep{sql-palm} and CodeS \citep{codes}, which enhanced models via secondary pre-training on large SQL corpora. BASE-SQL \citep{base-sql} also adopted a multi-stage fine-tuning framework. These SFT methods have proven to be effective pathways for improving Text-to-SQL performance on open-source models.

\begin{figure}[!htb]
    \centering

    \includegraphics[width=0.48\textwidth]{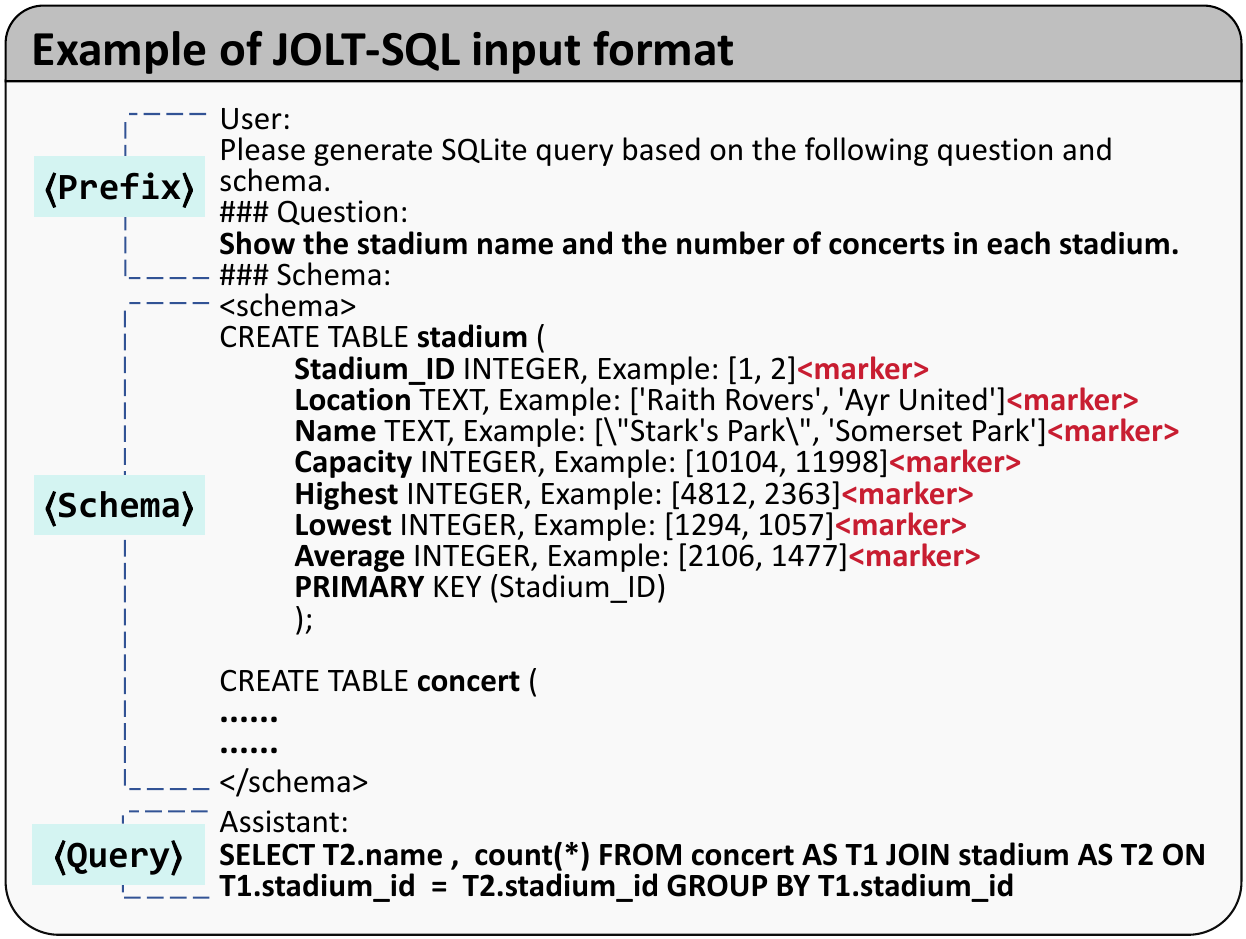}
    \caption{An input example of JOLT-SQL.
    }
    \label{schema_format}
\end{figure}

\section{Approach}
In this section, we detail our proposed JOLT-SQL method. We begin by describing its unique schema formatting, followed by an exploration of the joint loss design for schema linking and SQL generation tasks.

\subsection{Schema Representation Standardization}
We first standardize the schema representation to ensure a uniform input format for subsequent processing stages.

In JOLT-SQL, we divide the input sequence during the training phase into three distinct parts based on content: \( \langle\texttt{Prefix}\rangle\), \( \langle\texttt{Schema}\rangle\), and \( \langle\texttt{Query}\rangle\), as illustrated in Figure \ref{schema_format}. Specifically:

$\bullet$ \(\langle\texttt{Prefix}\rangle\): Contains essential task instructions and the user's question.

$\bullet$ \(\langle\texttt{Schema}\rangle\): The schema section, describing the complete table structure and column definitions.

$\bullet$ \(\langle\texttt{Query}\rangle\): The ground truth SQL query statement part of each input sequence, used for supervised training.

We employ a Data Definition Language (DDL)-style representation for the schema \citep{xiyan-sql}. For each table in the database, we specify the name and data type of every column, along with its primary and foreign key information. Additionally, for each column, we select at most two value examples from the database to include in its definition.

After each column definition within \(\langle\texttt{Schema}\rangle\), we insert a special marker token, denoted as \(\langle\texttt{marker}\rangle\). We simply select the model's padding token (e.g., "\texttt{<|endoftext|>}" from the Qwen2 tokenizer) as our marker token. As a special token, it acts as an isolated token that avoids combining with other characters into subwords, thereby simplifying subsequent processing, and is also semantically neutral.

\subsection{Discriminative Schema Linking with Local Bidirectional Attention}
\label{disSl}
Unlike typical schema linking methods based on generative LLMs, and inspired by LS-LLaMA \citep{ls-llama}, our approach relies on directly discriminating the hidden states of marker tokens.

Specifically, given a complete input sequence \(X=\langle\texttt{Prefix, Schema, Query}\rangle=[x_1, x_2, \dots, x_n]\), after it passes through the LLM decoder layers, we obtain its hidden states \(H \in \mathbb{R}^{n \times d}\). For the hidden state \(h_i\) corresponding to the $i$-th token, we pass it through a linear layer \(W \in \mathbb{R}^{1 \times d}\) and a sigmoid function to convert it into a probability:
\begin{equation}
\hat{y}_i=\sigma(Wh_i)
\label{sigmoid}
\end{equation}

Concurrently, for each input sequence \(X\), we create a binary mask \(M=[m_1, m_2, \dots, m_n]\). An element \(m_i\) is set to 1 if the \(i\)-th token is a marker token located within the \( \langle\texttt{Schema}\rangle \), and 0 otherwise. The loss is defined as:

{\footnotesize
\begin{equation}
\mathcal{L}_{\text{SL}} = -\frac{1}{\sum_{i=1}^{n}m_i} \sum_{i=1}^{n} m_i \cdot \text{BCE}(\hat{y}_i,y_i)
\end{equation}
}

Here, BCE means binary cross-entropy loss, \(y_i \in \{0,1\}\) is the ground truth label for the schema linking task. 

Furthermore, decoder-only LLM architectures employ a causal attention mask, which prevents tokens within \(\langle\texttt{Schema}\rangle\) from accessing global schema information. As an improvement, we enable tokens within the \(\langle\texttt{Schema}\rangle\) part to additionally have local bidirectional attention, on top of the default causal attention. Let \(A(x_i)\) denote the set of indices of other tokens that $x_i$ can attend to, and \({I}\) denote the set of indices for token positions in the sequence, \(A(x_i)\) is defined as:

{\footnotesize 
\begin{equation}
\begin{aligned} 
 \forall i \in & I_{\langle\texttt{Schema}\rangle}, \  A(x_i) = \\ 
& 
\begin{cases}
(I_{\langle\texttt{Prefix}\rangle} \cup I_{\langle\texttt{Schema}\rangle}) \setminus I_{\langle\texttt{marker}\rangle} , & i \notin I_{\langle\texttt{marker}\rangle} \\
I_{\langle\texttt{Prefix}\rangle} \cup I_{\langle\texttt{Schema}\rangle}, & i \in I_{\langle\texttt{marker}\rangle}
\end{cases}
\end{aligned}
\end{equation}
}

It should be noted that we introduce a further adjustment rule for marker tokens: marker tokens are invisible to all non-marker tokens. This modification allows us to remove marker tokens during the SQL  generation stage.

\subsection{SQL Supervised Tuning with Schema Selective Attention}
\label{sec:sa}
For the SQL query generation task, we enhance the model performance by applying supervised tuning to the \( \langle\texttt{Query}\rangle \) part. A challenge here is that during training, the input sequence contains the full \( \langle\texttt{Schema}\rangle \). For pipeline SFT methods, the SQL generation model is often fine-tuned on the ground truth schema subset, which helps to maintain consistency with the input conditions expected during inference \citep{dts-sql}.

We address this issue by similarly adjusting the attention mask. Specifically, tokens in the \(\langle\texttt{Query}\rangle\) part selectively attend only to tokens within the ground truth schema subset (referred to as \(\langle\texttt{GT\_Schema}\rangle\)), rather than the full \(\langle\texttt{Schema}\rangle\). In practice, \(\langle\texttt{GT\_Schema}\rangle\) includes not only the relevant column definitions but also the table structure to which each column definition belongs (such as table names, primary key, and foreign key definitions).

In addition to \(\langle\texttt{GT\_Schema}\rangle\), to make the model more robust to imperfect schema linking results, we select some noisy schema items (referred to as \(\langle\texttt{Noisy\_Schema}\rangle\)) for the \(\langle\texttt{Query}\rangle\) tokens to attend to. This attention mechanism can be represented as:

{\footnotesize
\begin{equation}
\begin{split}
 \forall i \in I_{\langle\texttt{Query}\rangle}, & \ A(x_i) = \\ 
 & \bigl( I_{\langle\texttt{Prefix}\rangle} \cup I_{\langle\texttt{GT\_Schema}\rangle} \cup I_{\langle\texttt{Noisy\_Schema}\rangle} \cup {} \\ 
                 & \{j | j \in I_{\langle\texttt{Query}\rangle} , j \leq i \} \bigr) \setminus I_{\langle\texttt{marker}\rangle}
\end{split}
\end{equation}
}

\begin{figure}[!htb]
\centering

\includegraphics[width=0.37\textwidth]{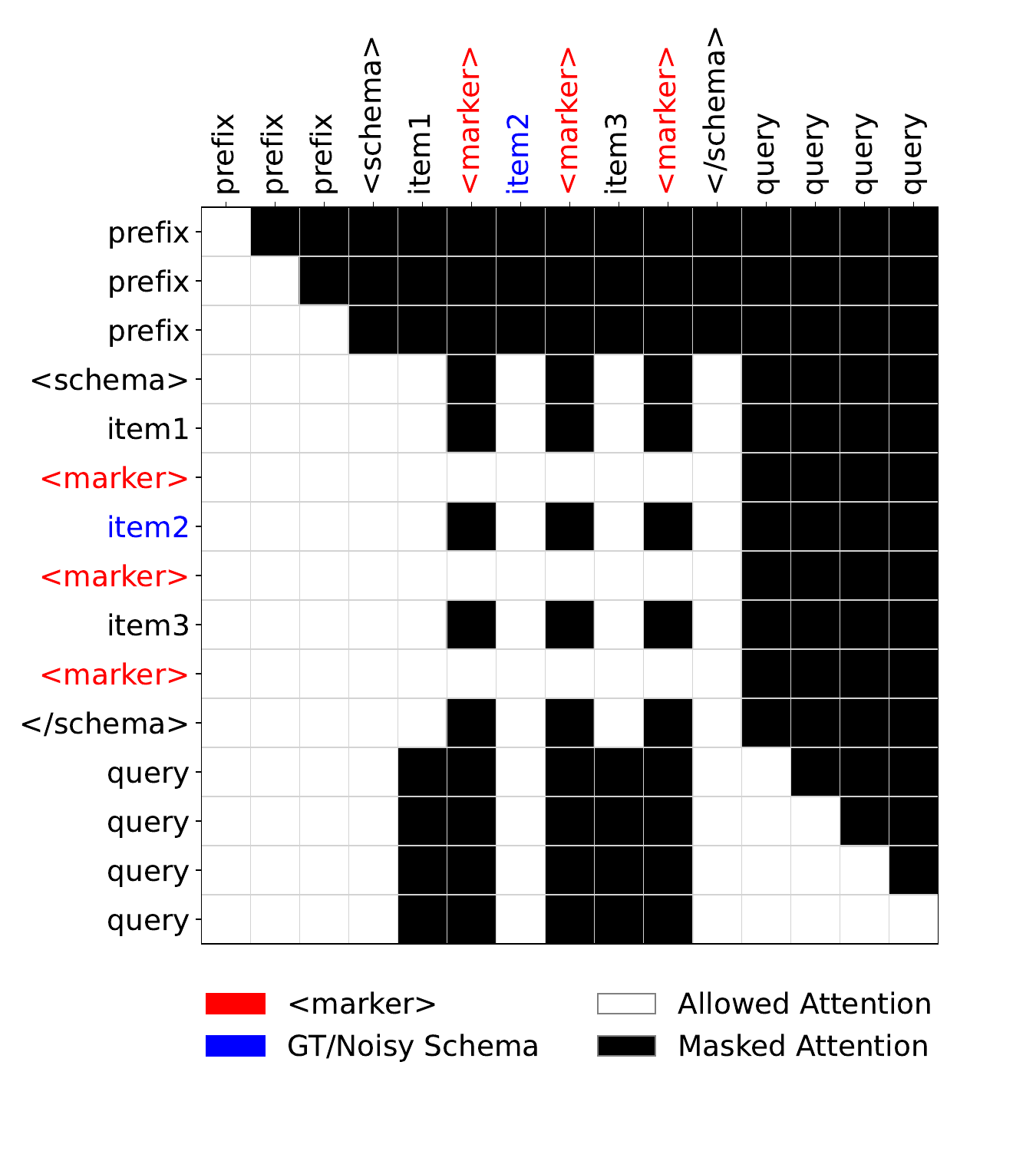}
\caption{Visualize the attention mask of JOLT-SQL using a sample text. For clarity, the text has been simplified and does not reflect the actual tokenization.}
\label{mask_vis}
\end{figure}

Combined with the schema Local Bidirectional Attention described in Section \ref{disSl}, the final attention mask form is shown in Figure \ref{mask_vis}.

Consistent with typical LLM SFT, we then calculate the Next Token Prediction (NTP) loss for the SQL query. For a \(\langle\texttt{Query}\rangle=[x_{n-m+1}, \dots, x_n]\), the loss is:

{\scriptsize
\begin{equation}
\begin{split}
\mathcal{L}_{\text{NTP}} = -\frac{1}{m}\sum_{i=n-m+1}^{n}\log P_{A(x_i)}\big( x_i \big| \langle \texttt{Prefix},\texttt{Schema} , x_{n-m+1:i-1}\rangle\big)
\end{split}
\end{equation}
}

where \(P_{A(\cdot)}(\cdot|\cdot)\) denotes the conditional probability given the application of the attention \(A(\cdot)\).

\begin{figure}[!htb]
    \centering

    \includegraphics[width=0.48\textwidth]{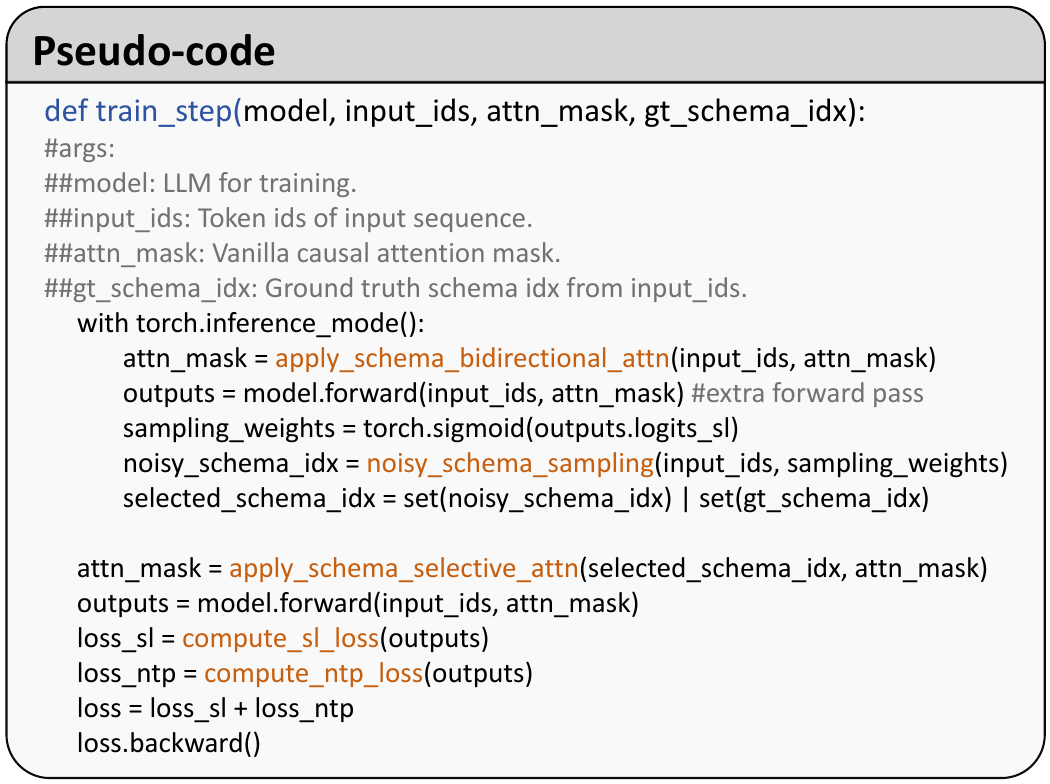}
    \caption{Pseudo-code for the JOLT-SQL training process. In the actual experiments, the \texttt{extra forward pass} is performed only during the first training epoch, and the \texttt{sampling\_weights} are cached.
    }
    \label{pseudo_code}
\end{figure}

\subsection{Joint Loss Tuning with Confusion-aware Noisy Schema Sampling}
\label{sec:joint_loss_tuning}

In the preceding sections, we have introduced the respective losses for the schema linking task \( \mathcal{L}_{\text{SL}}\) and the SQL generation task \(\mathcal{L}_{\text{NTP}}\). In JOLT-SQL, we train the model using a joint loss:

{\footnotesize
\begin{equation}
\mathcal{L}=\mathcal{L}_{\text{SL}}+\mathcal{L}_{\text{NTP}}
\label{eq:total_loss} 
\end{equation}
}

As described in Section \ref{sec:sa}, we introduce attention to \( \langle \texttt{Noisy\_Schema} \rangle \). Instead of simple random sampling, we employ a probability-weighted sampling to select these noisy schema items. The motivation here is that selecting noisy schema items that the model tends to misidentify with high confidence can help the model better adapt to schema linking results that may contain numerous False Positives during inference.

We define a func {\footnotesize\(Sample(Sets, Weights, Count)\)}, which performs sampling without replacement of \(Count\) items from the \(Sets\) based on their \(Weights\). Let \(\mathcal{S}\) denote the set of schema items, then:

\vspace{-4pt}
{\footnotesize
\begin{equation}
\begin{split}
 \mathcal{S}_{\langle \texttt{Noisy\_Schema}\rangle}= & Sample(\mathcal{S}_{\langle \texttt{Schema}\rangle} \setminus \mathcal{S}_{\langle \texttt{GT\_Schema}\rangle}, \ \hat{y},\  \lfloor k \rfloor ) \\
k \sim & \  U\big(0,\big \lfloor \beta \cdot |\mathcal{S}_{\langle \texttt{Schema}\rangle} | \big \rfloor \big)
\end{split}
\end{equation}
}
\vspace{-2pt}
Here, \(\beta\) is a float in the range (0,1) that controls the upper bound for the number of samples \(k\). In our experiments, \(\beta\) is set to 0.2 for the Spider dataset and 0.1 for the BIRD dataset.

Thanks to our joint loss tuning strategy, probability-weighted sampling can be efficiently integrated into the training process. During training, this probability-weighted sampling is guided by the model's own propensity for confusion: its predicted probabilities \(\hat{y}\) (Equation \ref{sigmoid}) for marker tokens are dynamically used as the sampling weights. These weights are obtained through an extra forward pass that does not require gradient computation. The pseudo-code for the training process is shown in Figure \ref{pseudo_code}.

Notably, these sampling weights are calculated and cached during the first training epoch and reused in subsequent epochs. The rationale for this strategy is our aim to capture the model's probability distribution when encountering "new" data, which more closely reflects its behavior in actual inference scenarios. In the first epoch, all training data is unseen by the model. In subsequent epochs, as the model progressively fits the training data, its prediction confidence on this seen data increases rapidly, causing the predicted probabilities to become extreme (approaching 0 or 1). This deviates from the model's behavior on unseen data (dev/test sets).

Furthermore, this caching strategy significantly improves training efficiency. The additional time cost incurred by the extra forward pass is minimal. For a more detailed comparison of time efficiency, please refer to Appendix \ref{latency_comp}.
\vspace{-1pt}
\subsection{Inference}
During schema linking inference, we prioritize recall by setting the decision threshold for \(\hat{y}\) to 0.05. This emphasis on high recall is crucial, as missing necessary schema items is more detrimental to subsequent SQL generation than including some redundant ones, which models can often partially handle. 

Following schema linking, we prune all marker tokens and irrelevant column definitions identified by the linking results. We then generate the SQL query using a vanilla causal attention mask. This ensures compatibility with frameworks such as HuggingFace Transformers' \texttt{model.generate()}, which do not support custom attention masks in autoregressive decoding.

\section{Experiments}
\subsection{Setup}

\paragraph{Datasets}
To evaluate the JOLT-SQL method, we conducted experiments on two widely used datasets: Spider \citep{spider} and BIRD \citep{bird}. \textbf{Spider} comprises 200 databases with multiple tables, and its Training/Dev/Test sets contain 7,000/1,034/2,147 question-SQL pairs, respectively. \textbf{BIRD} contains 12,751 unique question-SQL pairs, 95 large databases with a total size of 33.4 GB, and its Training/Dev/Test sets include 9,428/1,534/1,789 pairs, respectively. For these two benchmarks, our models are fine-tuned exclusively on their respective Training sets and evaluated on the corresponding Dev and Test sets.

\paragraph{Evaluation Metrics}
The quality of SQL generation is evaluated by Execution Accuracy (EX), which measures whether the execution results of the generated SQL on the database exactly match the ground truth answers. This directly reflects the model's practical ability to generate correct SQL.

For schema linking, we use Recall and Precision for evaluation. To more comprehensively evaluate the model's overall discriminative ability across different decision thresholds, we also introduce PR-AUC (Area Under the Precision-Recall Curve) and ROC-AUC (Area Under the Receiver Operating Characteristic Curve).

\paragraph{Model}
We selected the popular open-source Qwen2.5-Coder \citep{qwen2.5,qwen2.5-coder} series for our experiments, including Qwen2.5-Coder-7B-Instruct and Qwen2.5-Coder-14B-Instruct.

\paragraph{Implementation Details}
We employ LoRA \citep{lora} to reduce VRAM requirements. More details can be found in the Appendix \ref{app:training_details}.

\paragraph{Baselines}
In our experiments, we compare recent LLM-based Text-to-SQL baselines in two categories. The Prompting methods include DIN-SQL \citep{din-sql}, DAIL-SQL \citep{dail-sql}, CHESS \citep{chess}, E-SQL \citep{e-sql}, MCS-SQL \citep{mcs-sql}, RSL-SQL \citep{rsl-sql} and MAC-SQL \citep{mac-sql}. The SFT-based methods comprise CodeS \citep{codes}, SENSE \citep{sense}, DTS-SQL \citep{dts-sql}, ExSL \citep{exsl}, ROUTE \citep{route}, DB-Explore \citep{db-explore}, and BASE-SQL \citep{base-sql}. These baselines cover a range of prompting and fine-tuning strategies for Text-to-SQL.

\begin{table*}[!ht]

\centering
\resizebox{1.0\textwidth}{!}{
\begin{tabular}{lcccc}
\toprule[1pt]
\textbf{Methods}                            & \textbf{Type}      & \begin{tabular}[c]{@{}c@{}}\textbf{Spider Dev}\\ \textbf{EX}\end{tabular} & \begin{tabular}[c]{@{}c@{}}\textbf{Spider Test}\\ \textbf{EX}\end{tabular} & \begin{tabular}[c]{@{}c@{}}\textbf{BIRD Dev}\\ \textbf{EX}\end{tabular} \\ \hline
DIN-SQL + GPT-4\citep{din-sql}                     & Prompting & 82.8                                                    & 85.3                                                     & 50.7                                                  \\
DAIL-SQL + GPT-4\citep{dail-sql}                    & Prompting & 83.5                                                    & 86.6                                                     & 54.8                                                  \\
CHESS + Gemini1.5-Pro\citep{chess}                & Prompting & -                                                       & 87.2                                                     & 68.3                                                  \\
E-SQL + GPT-4o-mini\citep{e-sql}                  & Prompting & -                                                       & 74.8                                                     & 61.6                                                  \\
MAC-SQL + GPT-4\citep{mac-sql}                      & Prompting & 86.8                                                    & 82.8                                                     & 59.4                                                  \\
RSL-SQL + GPT-4o\citep{rsl-sql}                     & Prompting & -                                                       & 87.9                                                     & 67.2                                                  \\
MCS-SQL + GPT-4\citep{mcs-sql}                     & Prompting & 89.5                                                    & 89.6                                                     & 63.4                                                  \\ \hline
CodeS-7B\citep{codes}                           & SFT       & 85.4                                                    & 83.5                                                        & 57.2                                                  \\
CodeS-15B\citep{codes}                          & SFT       & 84.9                                                    & 85.0                                                        & 58.5                                                  \\
SENSE-7B\citep{sense}                           & SFT       & 83.2                                                    & 83.5                                                     & 51.8                                                  \\
SENSE-13B\citep{sense}                          & SFT       & 84.1                                                    & 86.6                                                     & 55.5                                                  \\
DTS-SQL + Deepseek-7B\citep{dts-sql}                & SFT       & 85.5                                                    & 84.4                                                     & 55.8                                                  \\
ExSL + Deepseek-Coder-6.7B\citep{exsl}           & SFT       & 82.4                                                    & 83.0                                                     & 63.2                                                  \\
ROUTE + Qwen2.5-7B\citep{route}                 & SFT       & 83.6                                                    & 83.7                                                     & 55.9                                                  \\
ROUTE + Qwen2.5-14B\citep{route}                & SFT       & {\ul 87.3}                                              & 87.1                                                     & 60.9                                                  \\
DB-Explore + Qwen2.5-Coder-7B\citep{db-explore}      & SFT       & 84.0                                                    & -                                                        & 52.1                                                  \\
BASE-SQL + Qwen2.5-Coder-14B\citep{base-sql}       & SFT       & 86.8                                                    & {\ul 87.9}                                               & {\ul 63.8}                                            \\
JOLT-SQL(\textbf{Ours}) + Qwen2.5-Coder-7B  & SFT       & 87.0                                                    & 86.8                                                     & 60.4                                                  \\
JOLT-SQL(\textbf{Ours}) + Qwen2.5-Coder-14B & SFT       & \textbf{88.4}                                           & \textbf{88.9}                                            & \textbf{64.9}                                         \\ 
\bottomrule[1 pt]
\end{tabular}
}
\caption{Main experimental results. The best results among SFT-based methods are shown in \textbf{bold}, and the second-best are {\ul underlined}.}
\label{main_table}
	
\end{table*}

\begin{table*}[!ht]
\vspace{-0.1pt}
\centering
\resizebox{1.0\textwidth}{!}{
\begin{tabular}{l|cccc|cccc|cccc}
\toprule[1pt]
\multirow{2}{*}{\textbf{Methods}}  & \multicolumn{4}{c|}{\textbf{Spider Dev}}                          & \multicolumn{4}{c|}{\textbf{Spider Test}}                         & \multicolumn{4}{c}{\textbf{BIRD Dev}}                             \\
                                   & \textbf{P}     & \textbf{R}     & \textbf{ROC}   & \textbf{PR}    & \textbf{P}     & \textbf{R}     & \textbf{ROC}   & \textbf{PR}    & \textbf{P}     & \textbf{R}     & \textbf{ROC}   & \textbf{PR}    \\ \hline
ExSL + Deepseek-Coder-6.7B\citep{exsl}           & -              & -              & 99.76          & 98.44          & -              & -              & 99.70          & 98.40          & -              & -              & 99.38          & 93.67          \\
JOLT-SQL(\textbf{Ours}) + Qwen2.5-Coder-7B  & 88.09          & 98.12          & 99.86          & 98.70          & 90.28          & \textbf{98.65} & 99.87          & 99.10          & \textbf{76.14} & 94.83          & 99.46          & 94.31          \\
JOLT-SQL(\textbf{Ours}) + Qwen2.5-Coder-14B & \textbf{91.07} & \textbf{99.03} & \textbf{99.91} & \textbf{99.29} & \textbf{93.74} & 98.58          & \textbf{99.89} & \textbf{99.26} & 75.12          & \textbf{96.68} & \textbf{99.54} & \textbf{95.42} \\ \bottomrule[1 pt]
\end{tabular}
}
\caption{Schema linking results. \textbf{P} and \textbf{R} represent column Precision and column Recall, respectively, at a decision threshold of 0.05. \textbf{ROC} stands for ROC AUC, and \textbf{PR} stands for PR AUC. The best results are shown in \textbf{bold}.}
\label{sl_table}
\vspace{-2pt}	
\end{table*}

\vspace{-1pt}
\subsection{Main Results}
Table \ref{main_table} presents our main experimental results and a comparison with baselines. JOLT-SQL demonstrates excellent performance on both the Spider and BIRD benchmarks. Specifically, JOLT-SQL using the Qwen2.5-Coder-14B model achieves execution accuracies of 88.4\%, 88.9\%, and 64.9\% on the Spider Dev set, Test set, and BIRD Dev Set, respectively. This performance is state-of-the-art among open-source models of comparable size based on SFT. Notably, JOLT-SQL with Qwen2.5-Coder-7B even surpasses larger models like CodeS-15B and SENSE-13B in results.

Unlike the two-stage fine-tuning strategy of DTS-SQL or the complex multi-stage pipeline of BASE-SQL, JOLT-SQL achieves superior performance and significantly simplifies the training process through its novel single-stage joint loss tuning. Meanwhile, compared to the multi-task learning approach adopted by ROUTE, JOLT-SQL achieves better results while keeping the total amount of training data unchanged. Furthermore, even when compared to prompting methods that often rely on powerful closed-source models like GPT-4 \citep{gpt4,gpt4o} or Gemini1.5-Pro \citep{gemini1.5}, JOLT-SQL demonstrates strong competitiveness, with its performance matching or exceeding some of these top approaches.

\begin{figure}[!hbt]

\centering
\subfloat[Spider Dev]{
\includegraphics[scale=0.23]{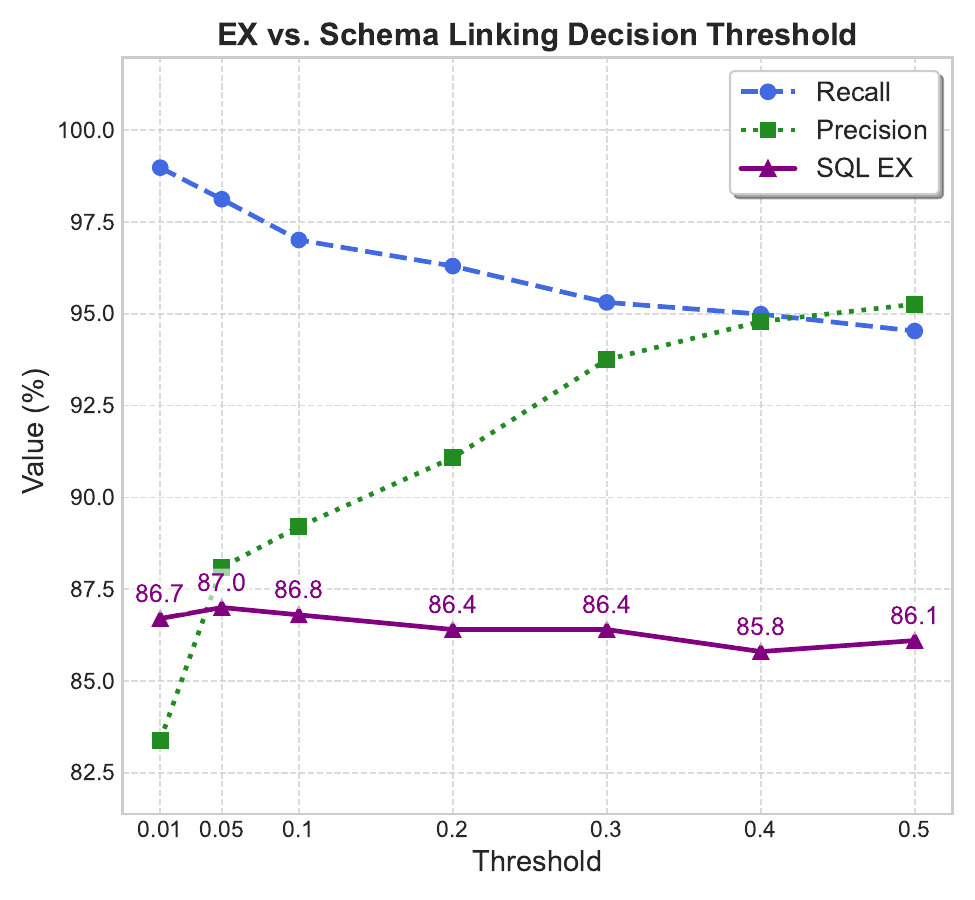}}
\hspace{-3pt}
\subfloat[BIRD Dev]{
\includegraphics[scale=0.23]{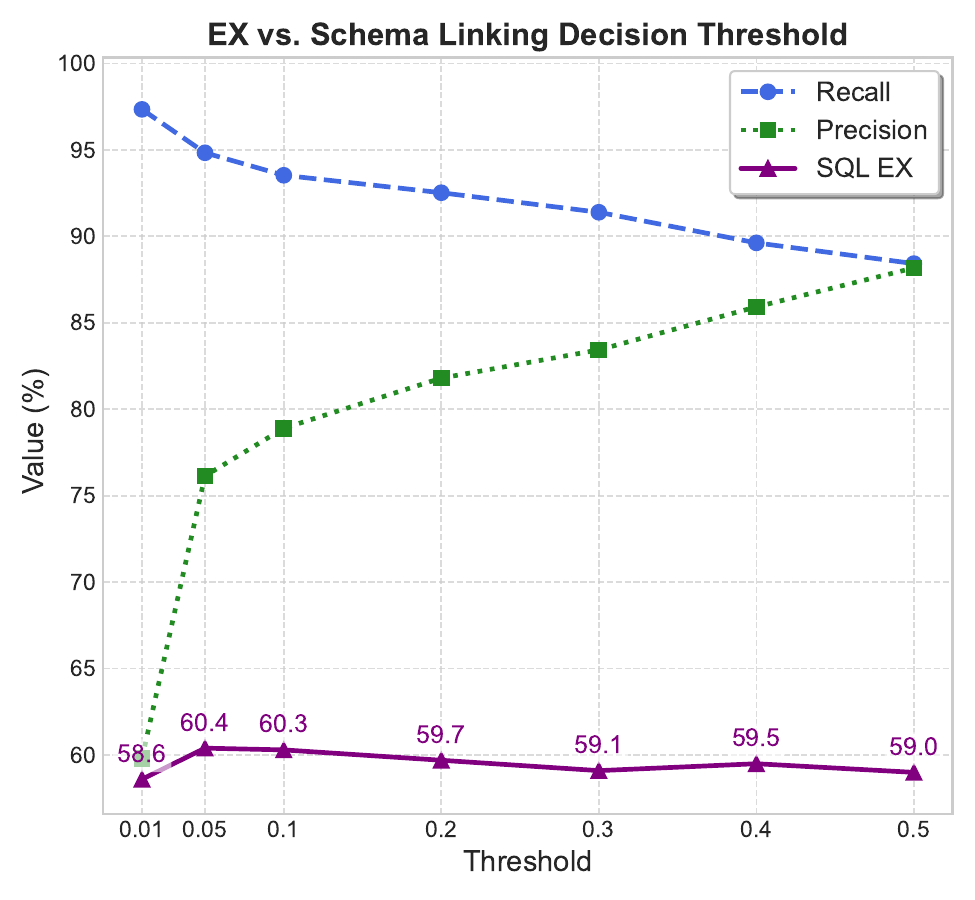}}

\caption{Comparison of the model's final EX under different schema linking decision thresholds.}
\label{plot_threshold}

\end{figure}

\subsection{Schema Linking Results}
Table \ref{sl_table} displays the schema linking task metrics of our JOLT-SQL method on the Spider Dev/Test and BIRD Dev sets. We compare it with ExSL \citep{exsl} as a baseline. ExSL employs an extractive method to fine-tune decoder-only LLMs and relies on repeating column definitions to alleviate the issue of incomplete schema information caused by causal masks. In contrast, JOLT-SQL, by introducing a local bidirectional attention mechanism, avoids the need to redundantly duplicate column definitions, which would otherwise increase sequence length. The results show that when using the Qwen2.5-Coder-14B model, JOLT-SQL achieves ROC and PR scores as high as 99.91\% and 99.29\%, respectively, on the Spider Dev set, and also attains SOTA performance of 99.54\% ROC and 95.42\% PR on the BIRD Dev set.

We further investigated the impact of the schema linking threshold on the specific performance metrics of JOLT-SQL (based on the Qwen2.5-Coder-7B). In Figure \ref{plot_threshold}, we present the changes in column recall, column precision, and the final SQL EX as the decision threshold decreases from 0.5 to 0.01. On the Spider Dev set, it can be observed that as the decision threshold starts to decrease from 0.5, the EX shows an overall upward trend, reaching its peak EX at a threshold of 0.05. Opting for an even lower threshold of 0.01 results in only a slight improvement in recall, accompanied by a significant sacrifice in precision. This leads to the introduction of excessive False Positive schema items during inference, causing the EX to decline. 

BIRD exhibits a similar overall trend with respect to different thresholds, reaching its peak EX around 0.05. Notably, compared to the Spider dataset, an increase in column recall on the BIRD dataset is typically accompanied by a more rapid decline in precision. This might reflect the model's higher uncertainty when dealing with a more complex dataset like BIRD.

\subsection{Ablation Study}
In the ablation study, we analyzed the impact of three core components on overall performance: schema Local Bidirectional Attention \textbf{LBA}, Confusion-aware Noisy Schema Sampling (\textbf{Confusion-aware NSS}) and schema Selective Attention \textbf{SA}. In the context of these ablations, SA defaults to the \(\langle\texttt{Query}\rangle\) attending only to the \(\langle\texttt{GT\_Schema}\rangle\) part of the overall schema information (thus excluding \(\langle\texttt{Noisy\_Schema}\rangle\), which would be introduced by the NSS), with NSS treated as an additional component.

The results of the ablation study, detailed in Table \ref{ablation_table}, confirm that all three core components contribute to JOLT-SQL's performance. LBA was vital for capturing complex schema relationships, and its absence markedly impacted results. NSS proved key for improving robustness against noisy schema during inference by promoting adaptation in training, while SA helped align training with inference conditions by focusing \(\langle\texttt{Query}\rangle\) attention on relevant schema subsets instead of the full schema. The individual or combined removal of these mechanisms consistently reduced performance. Ultimately, disabling all three components substantially degraded overall performance.

The results of the ablation study demonstrate that within our joint loss tuning strategy, the clever integration of these attention adjustment and noise injection mechanisms plays an indispensable role in achieving superior model performance.

\begin{table}[!htb]
\centering
\resizebox{1.0\linewidth}{!}{
\begin{tabular}{lcc}
\toprule[1pt]
\textbf{Method}                          & \textbf{\begin{tabular}[c]{@{}c@{}}Spider\\ Dev EX\end{tabular}} & \textbf{\begin{tabular}[c]{@{}c@{}}BIRD\\ Dev EX\end{tabular}} \\ \hline
JOLT-SQL + Qwen2.5-Coder-7B              & \textbf{87.0}                                                    & \textbf{60.4}                                                  \\
\textit{w/o} LBA                                & 84.8                                                             & 58.3                                                           \\
\textit{w/o} Confusion-aware NSS              & 86.1                                                             & 58.6                                                           \\
\textit{w/o} SA \& Confusion-aware NSS        & 85.9                                                             & 59.2                                                           \\
\textit{w/o} LBA \& SA \& Confusion-aware NSS & 84.5                                                             & 57.7                                                           \\ \bottomrule[1 pt]
\end{tabular}
}
\caption{Ablation study results.}
\label{ablation_table}
	
\end{table}

\begin{table*}[!ht]

\centering
\resizebox{0.8\textwidth}{!}{
\begin{tabular}{l|cccc|cccc}
\toprule[1pt]
\multirow{2}{*}{\textbf{Methods}} & \multicolumn{4}{c|}{\textbf{Spider Dev}}                          & \multicolumn{4}{c}{\textbf{BIRD Dev}}                             \\
                                  & \textbf{P}     & \textbf{R}     & \textbf{ROC}   & \textbf{PR}    & \textbf{P}     & \textbf{R}     & \textbf{ROC}   & \textbf{PR}    \\ \hline
JOLT-SQL + Qwen2.5-Coder-7B       & \textbf{88.09} & \textbf{98.12} & \textbf{99.86} & \textbf{98.70} & \textbf{76.14} & \textbf{94.83} & \textbf{99.46} & \textbf{94.31} \\
\textit{w/o} LBA                   & 79.83          & 95.04          & 99.34          & 95.41          & 66.01          & 92.73          & 98.86          & 88.02          \\ \bottomrule[1 pt]
\end{tabular}
}
\caption{Ablation results for schema LBA.}
\label{sl_bi_attn_ablation}

\end{table*}

\subsection{Further Analysis}

\begin{table}[!htb]
\centering
\resizebox{1.0\linewidth}{!}{
\begin{tabular}{l|ll|cc}
\toprule[1pt]
\textbf{No.} & \textbf{Training with(\(\langle\texttt{Query}\rangle\) attends to)}             & \textbf{Inference with}    & \textbf{\begin{tabular}[c]{@{}c@{}}Spider \\ Dev EX\end{tabular}} & \textbf{\begin{tabular}[c]{@{}c@{}}BIRD\\ Dev EX\end{tabular}} \\ \hline
\textbf{\#1} & Full \(\langle\texttt{Schema}\rangle\)                         & Full \(\langle\texttt{Schema}\rangle\)                 & 82.8                                                              & 52.9                                                           \\ \hline
\textbf{\#2} & Full \(\langle\texttt{Schema}\rangle\)                         & \(\langle\texttt{GT\_Schema}\rangle\)                  & 89.6                                                              & 68.5                                                           \\
\textbf{\#3} & \(\langle\texttt{GT\_Schema}\rangle\)                         & \(\langle\texttt{GT\_Schema}\rangle\)                 & \textbf{90.5}                                                     & \textbf{72.2}                                                  \\ \hline
\textbf{\#4} & Full \(\langle\texttt{Schema}\rangle\)                        & \(\hat{y}>0.05\) \texttt{Schema} & 85.9                                                              & 59.2                                                           \\
\textbf{\#5} & \(\langle\texttt{GT\_Schema}\rangle\)                          & \(\hat{y}>0.05\) \texttt{Schema} & 86.1                                                              & 58.6                                                           \\
\textbf{\#6} & \(\langle\texttt{GT\_Schema}\rangle\) \& Random NSS            & \(\hat{y}>0.05\) \texttt{Schema} & 86.4                                                              & 59.6                                                           \\
\textbf{\#7} & \(\langle\texttt{GT\_Schema}\rangle\) \& Confusion-aware NSS & \(\hat{y}>0.05\) \texttt{Schema} & {\ul 87.0}                                                        & {\ul 60.4}                                                     \\ \bottomrule[1pt]
\end{tabular}
}
\caption{SQL generation performance under different training attention strategies and inference-stage schema input combinations (based on Qwen2.5-Coder-7B, with LBA enabled). \(\hat{y}>0.05\) \texttt{Schema} refers to the results predicted by schema linking. \textbf{\#1}, \textbf{\#2}, and \textbf{\#3} are used for baseline or ideal condition comparisons, with \textbf{\#3} considered as a performance upper bound. \textbf{\#4} to \textbf{\#7} focus on the model's performance under realistic predicted schema conditions, where \textbf{\#7} represents the complete strategy adopted by JOLT-SQL.}
\label{schema_comp}
\end{table}

\paragraph{Impact of Schema LBA}
The ablation study has already shown that removing LBA leads to a significant decline in model performance. Table \ref{sl_bi_attn_ablation} further reveals its critical role in schema linking: the absence of LBA causes a deterioration in both recall and precision, posing a dual challenge for SQL generation due to increased noise and omission of relevant columns.

For instance, on the Spider Dev/BIRD Dev sets, after removing LBA, even with a low decision threshold of 0.05, recall dropped to 95.04\% and 92.73\%, respectively, while precision significantly decreased to 79.83\% and 66.01\%. The PR also correspondingly decreased from 98.70\%/94.31\% to 95.41\%/88.02\%,  confirming the weakened ability of the model to identify relevant columns. This is primarily because, under the constraints of the standard causal mask, most marker tokens can only perceive local preceding information and cannot make judgments by synthesizing complete schema information, thus highlighting the necessity of LBA.

\paragraph{Impact of Schema SA}
The experimental results in Table \ref{schema_comp} reveal the role of Schema SA during the training phase. Comparing \textbf{\#3} (using \(\langle\texttt{GT\_Schema}\rangle\) for both training and inference) with \textbf{\#1} (using Full \(\langle\texttt{Schema}\rangle\) for both), the former, serving as an ideal performance upper bound, significantly outperforms the latter, indicating the necessity of schema linking. \textbf{\#2} (training with Full \(\langle\texttt{Schema}\rangle\) attention, inference with \(\langle\texttt{GT\_Schema}\rangle\) input) shows significant improvement over \textbf{\#1}, but still falls short of the ideal upper bound of \textbf{\#3}. This highlights the importance of consistency in input distribution between the training and inference phases. Schema SA achieves this input alignment between training and inference phases by cleverly defining the attention scope.

\paragraph{Impact of NSS}
In more realistic inference scenarios, the model must effectively handle potentially imperfect schema subsets obtained from the schema linking phase. Experiments \textbf{\#4} to \textbf{\#7} in Table \ref{schema_comp} explore this aspect. The EX of \textbf{\#4} (which corresponds to the "\textit{w/o} SA \& Confusion-aware NSS" setting in Table \ref{ablation_table}) can be considered a baseline performance. Subsequently, the EX of \textbf{\#5} (also the "\textit{w/o} Confusion-aware NSS" setting in Table \ref{ablation_table}) slightly increased to 86.1\% on Spider but dropped to 58.6\% on BIRD. This result indicates that although focusing the model on the ideal \(\langle\texttt{GT\_Schema}\rangle\) via SA during training is beneficial, a lack of adaptive training for noise still leaves the model insufficiently robust when facing imperfectly predicted schema.

To enhance the model's ability to handle such noise, we introduced the NSS strategy. Experiment \textbf{\#6} (training with \(\langle\texttt{GT\_Schema}\rangle\) + random NSS), compared to \textbf{\#5}, introduced randomly selected noisy schema items during training, boosting EX on Spider and BIRD to 86.4\% and 59.6\%, respectively. This preliminarily demonstrates the importance of exposing the model to and training it to handle noise.

Building on this, \textbf{\#7}, which is the strategy adopted by JOLT-SQL, achieved the best results. \textbf{\#7} follows the same training procedure as \textbf{\#6}—both attend to \(\langle\texttt{GT\_Schema}\rangle\) while introducing noisy items—but the key difference lies in the selection method for these noisy items: \textbf{\#6} uses simple random sampling, whereas the confusion-aware approach in \textbf{\#7} is reflected in its weighted sampling based on the model's predicted probabilities in the training phase. This sampling mechanism tends to prioritize items that the model erroneously assigns a high relevance probability to (potential False Positives). Considering that we intentionally use a low decision threshold during inference to boost recall (which introduces more False Positives), this sampling strategy, focused on learning to handle "confusing" items, is particularly well-suited. By specifically training on schema items prone to model misjudgment, JOLT-SQL can more effectively learn to discern and ignore the most disruptive redundant information. A case study on NSS is provided in Appendix \ref{sec:appendix_case_study}.

\paragraph{More Comparison \& Efficiency Analysis}
Beyond improving execution accuracy, JOLT-SQL also demonstrates superior training and inference efficiency compared to pipeline SFT methods. See Appendix \ref{latency_comp} for details.

\section{Conclusion}
In this paper, we introduced JOLT-SQL, an innovative SFT framework for Text-to-SQL. JOLT-SQL features a novel joint loss tuning method and incorporates confusion-aware noisy schema sampling to enhance model robustness. Experimental results compellingly demonstrate that JOLT-SQL achieves excellent SQL execution accuracy and exhibits clear advantages in efficiency. We believe this work offers new perspectives for developing more efficient and robust Text-to-SQL systems.

\section*{Limitations}

Despite the promising results, this work has several limitations. Firstly, our method was validated on relatively small-scale LLMs (7B and 14B parameters). Due to constraints in hardware resources and time, we were unable to conduct experiments to verify its effectiveness on larger models (e.g., Qwen2.5-Coder-32B). Secondly, while our experiments demonstrate JOLT-SQL's excellent efficiency during both training and inference, the method involves sophisticated adjustments to attention masks and custom logic integrated into the training iterations, particularly for dynamic attention mask generation and the noisy schema sampling process. This can introduce a degree of complexity at the code implementation level. 



\bibliography{arxiv}

\appendix
\section*{Appendix}

\section{Further Analysis}
\subsection{Comparison with Pipeline Fine-tuning Strategies \& Efficiency Analysis}
\label{latency_comp}

To further demonstrate the advantages of JOLT-SQL, this section compares it with DTS-SQL \citep{dts-sql}, a typical pipeline fine-tuning method. DTS-SQL employs a two-stage pipeline: first, fine-tuning an LLM for generative schema linking, and then fine-tuning another LLM for SQL generation based on ground truth schema input (table-level). We re-evaluated DTS-SQL using experimental settings consistent with those for JOLT-SQL. To align with our experimental setup, the SQL generation fine-tuning phase for DTS-SQL was based on column-level ground truth schema input. We compared their training efficiency, inference speed, and task performance, as shown in Table \ref{pipeline_comp_1}.

\textbf{In terms of training cost}, the results show that JOLT-SQL requires 5 hours and 5 minutes to complete 3 epochs of training. This duration includes an approximate 8.5\% overhead from enabling Confusion-aware NSS compared to its own baseline of 4 hours and 41 minutes (JOLT-SQL without Confusion-aware NSS), a time comparable to the standard SFT approach. Even so, JOLT-SQL is significantly more efficient than DTS-SQL. DTS-SQL, requiring independent fine-tuning for two stages, has a total training time of 7 hours and 10 minutes, which is 52.9\% longer than the JOLT-SQL baseline (without Confusion-aware NSS), highlighting the efficiency of JOLT-SQL's single-stage joint optimization. This is because JOLT-SQL learns both tasks from a single input sequence, fundamentally halving the total number of forward/backward passes compared to two-stage methods that process separate data for each task.

\textbf{In terms of inference speed}, JOLT-SQL's discriminative schema linking shows a significant advantage, averaging 0.11 seconds, which is much faster than DTS-SQL's generative linking at 0.57 seconds—the latter being more than five times slower. DTS-SQL's schema linking is more time-consuming mainly because it requires autoregressively generating relevant table and column names token by token during inference. In contrast, JOLT-SQL's discriminative method requires only a single full forward pass to determine the relevance of all schema candidates. Thanks to its efficient schema linking, JOLT-SQL's average end-to-end inference time is only 0.88 seconds, whereas DTS-SQL takes 1.34 seconds (approximately 52.3\% slower), with the difference primarily stemming from the schema linking stage.

\textbf{In terms of Execution Accuracy (EX)}, JOLT-SQL (87.0\%) outperforms DTS-SQL (84.8\%) on Spider Dev. This difference may stem from the characteristics of their schema linking approaches and the robustness of their SQL generation stages, as further elaborated by Table \ref{pipeline_comp_2}.

Table \ref{pipeline_comp_2} compares the schema linking metrics. DTS-SQL's generative linking achieves high precision 93.67\%, with a recall of 94.15\%. However, as discussed in the main text, JOLT-SQL's discriminative method can flexibly use a low decision threshold of 0.05 to prioritize recall, achieving a recall rate as high as 98.12\%. In Text-to-SQL tasks, high recall is crucial: omitting necessary items often leads to SQL failure, whereas a few redundant items have less impact if the SQL model is robust. The insufficient recall of DTS-SQL's schema linking might be one of the core reasons for its lower EX. Additionally, its SQL generation model is fine-tuned only on ground-truth relevant columns, lacking adaptability to imperfect inputs from actual schema linking.

The results again demonstrate that JOLT-SQL's joint loss tuning strategy is significantly superior to DTS-SQL's two-stage method in both training and inference efficiency. Its combination of discriminative schema linking and Confusion-aware NSS achieves better recall and model robustness, ultimately improving SQL EX.

\begin{table*}[!htb]
\centering

\resizebox{1.0\linewidth}{!}{
\begin{tabular}{lccccc}
\toprule[1 pt]
\textbf{Methods}                      & \textbf{\begin{tabular}[c]{@{}c@{}}Total \\ Training Time↓\end{tabular}} & \textbf{\begin{tabular}[c]{@{}c@{}}Avg. Schema Linking\\ Inference Time↓\end{tabular}} & \textbf{\begin{tabular}[c]{@{}c@{}}Avg. SQL Generation\\ Inference Time↓\end{tabular}} & \textbf{\begin{tabular}[c]{@{}c@{}}Avg. End-to-End\\ Inference Time↓\end{tabular}} & \textbf{\begin{tabular}[c]{@{}c@{}}Spider\\ Dev EX\end{tabular}} \\ \hline
Qwen2.5-Coder-7B + SFT                & 4h38min                                                                  & -                                                                                      & 0.94s                                                                                  & 0.94s                                                                              & 82.7                                                             \\ \hline
DTS-SQL\citep{dts-sql} + Qwen2.5-Coder-7B            & 7h10min(\textcolor{red}{+52.9\%})                                                         & 0.57s(\textcolor{red}{+418.2\%})                                                                        & 0.77s                                                                                  & 1.34s(\textcolor{red}{+54.0\%})                                                                     & 84.8                                                             \\
JOLT-SQL(\textbf{Ours}) + Qwen2.5-Coder-7B     & 5h5min(\textcolor{red}{+8.5\%})                                                           & \textbf{0.11s}                                                                         & \textbf{0.77s}                                                                         & \textbf{0.88s}                                                                     & \textbf{87.0}                                                    \\
\textit{w/o} Confusion-aware NSS(\textbf{Baseline}) & \textbf{4h41min}                                                         & \textbf{0.11s}                                                                         & \textbf{0.76s}                                                                         & \textbf{0.87s}                                                                     & 86.1                                                             \\ \bottomrule[1 pt]
\end{tabular}
}
\caption{Comparison results with the two-stage pipeline fine-tuning method DTS-SQL and a standard SFT approach (Qwen2.5-Coder-7B + SFT) are presented, all evaluated on a single NVIDIA A30 GPU. KV cache was enabled for all inference processes. We use JOLT-SQL without Confusion-aware NSS as the baseline for comparing training and inference times on the Spider Dev set. Training time is the total for 3 epochs. Inference time is the average per instance with Batchsize=1. For context, the cost for a single NVIDIA A30 instance was approximately \$0.18 per hour at the time of the experiments.
{\small  \textbf{Additional Notes:} \(\bullet\) The slightly longer SQL generation inference time for the standard SFT approach (Qwen2.5-Coder-7B + SFT) is due to its processing of the full schema, which results in longer input sequences. 
\(\bullet\) It's worth noting the distinction in forward pass latency between training and inference: our use of LoRA means adapters introduce some overhead during the training's forward pass, but for inference, these LoRA weights are merged into the base model, eliminating this adapter-induced latency.} }
\label{pipeline_comp_1}
	
\end{table*}

\begin{table*}[!htb]
\centering
\resizebox{0.8\textwidth}{!}{
\begin{tabular}{l|cccc}
\toprule[1 pt]
\multirow{2}{*}{\textbf{Methods}} & \multicolumn{4}{c}{\textbf{Spider Dev}}                           \\
                                  & \textbf{P}     & \textbf{R}     & \textbf{ROC}   & \textbf{PR}    \\ \hline
DTS-SQL\citep{dts-sql} + Qwen2.5-Coder-7B        & \textbf{93.67} & 94.15          & N/A            & N/A            \\
JOLT-SQL\textbf{(Ours)} + Qwen2.5-Coder-7B       & 88.09          & \textbf{98.12} & \textbf{99.86} & \textbf{98.70} \\ \bottomrule[1 pt]
\end{tabular}
}
\caption{Comparison of schema linking results between DTS-SQL's method (i.e., generative schema linking based on LLM SFT) and JOLT-SQL's discriminative method. \textbf{P} and \textbf{R} represent column Precision and column Recall, respectively. Notably, since generative LLMs output discrete schema linking results directly, probability information is not readily available, making standard ROC AUC and PR AUC metrics inapplicable.}
\label{pipeline_comp_2}
	
\end{table*}

\subsection{Cross-Dataset Generalization}

\begin{table*}[!htb]
\centering
\resizebox{0.9\textwidth}{!}{
\begin{tabular}{lcccc}
\toprule[1pt]
               \textbf{Methods}            & \textbf{Spider-Syn} & \textbf{Spider-Realistic} & \textbf{Spider-DK} & \textbf{KaggleDBQA} \\ \hline
JOLT-SQL + Qwen2.5-Coder-7B  & 77.2                & 83.0                      & 75.9               & 47.0                \\
JOLT-SQL + Qwen2.5-Coder-14B & 79.6                & 84.1                      & 79.1               & 52.4                \\ \bottomrule[1pt]
\end{tabular}
}
\caption{Execution Accuracy of JOLT-SQL on Spider variants and KaggleDBQA. The models were fine-tuned only on the original Spider training set.}
\label{generalization}
	
\end{table*}

Table \ref{generalization} presents JOLT-SQL's robustness and generalization capabilities across various challenging datasets. We evaluated its performance on Spider-Syn \citep{spider-syn}, Spider-Realistic \citep{spider-realistic}, Spider-DK \citep{spider-dk}, and the cross-domain KaggleDBQA dataset \citep{kaggledbqa}. The results indicate that JOLT-SQL maintains strong performance. Furthermore, it demonstrates solid generalization to entirely unseen databases and query types in KaggleDBQA.

\subsection{Performance in Low-Resource Settings}
\begin{table}[!htb]
\centering
\resizebox{0.45\textwidth}{!}{
\begin{tabular}{lcc}
\toprule[1pt]
\textbf{Training data used} & \textbf{\begin{tabular}[c]{@{}c@{}}Spider \\ Dev EX\end{tabular}} & \textbf{\begin{tabular}[c]{@{}c@{}}BIRD \\ Dev EX\end{tabular}} \\ \hline
5\%                         & 81.9                                                              & 54.0                                                            \\
10\%                        & 83.8                                                              & 55.3                                                            \\
25\%                        & 85.4                                                              & 58.5                                                            \\
50\%                        & 86.4                                                              & 59.7                                                            \\
100\%                       & \textbf{87.0}                                                     & \textbf{60.4}                                                   \\ \bottomrule[1pt]
\end{tabular}
}
\caption{Ablation results for JOLT-SQL under reduced data conditions (based on Qwen2.5-Coder-7B).}
\label{reduced_data}
	
\end{table}

Table \ref{reduced_data} evaluates JOLT-SQL's performance under low-resource training conditions. We fine-tuned the Qwen2.5-Coder-7B model using various proportions of the Spider and BIRD training set. The results indicate that the model's execution accuracy degrades gracefully as training data decreases, without a sharp drop. Even with only 5\% or 10\% of the data, JOLT-SQL still achieves acceptable performance, demonstrating its robustness and practical applicability in scenarios with limited annotated data.

\section{More Details}
\subsection{Training Data Construction}
We use SQLGlot\footnote{https://github.com/tobymao/sqlglot} to process SQL queries. By parsing SQL query statements to construct their Abstract Syntax Trees (ASTs) and leveraging SQLGlot's scope analysis feature, we can accurately identify each column referenced in the query and its originating data table, while also resolving table aliases. This ultimately allows us to extract ground truth schema linking pairs in the \texttt{table.column} format.

\begin{figure}[!htb]
\centering

\includegraphics[width=0.5\textwidth]{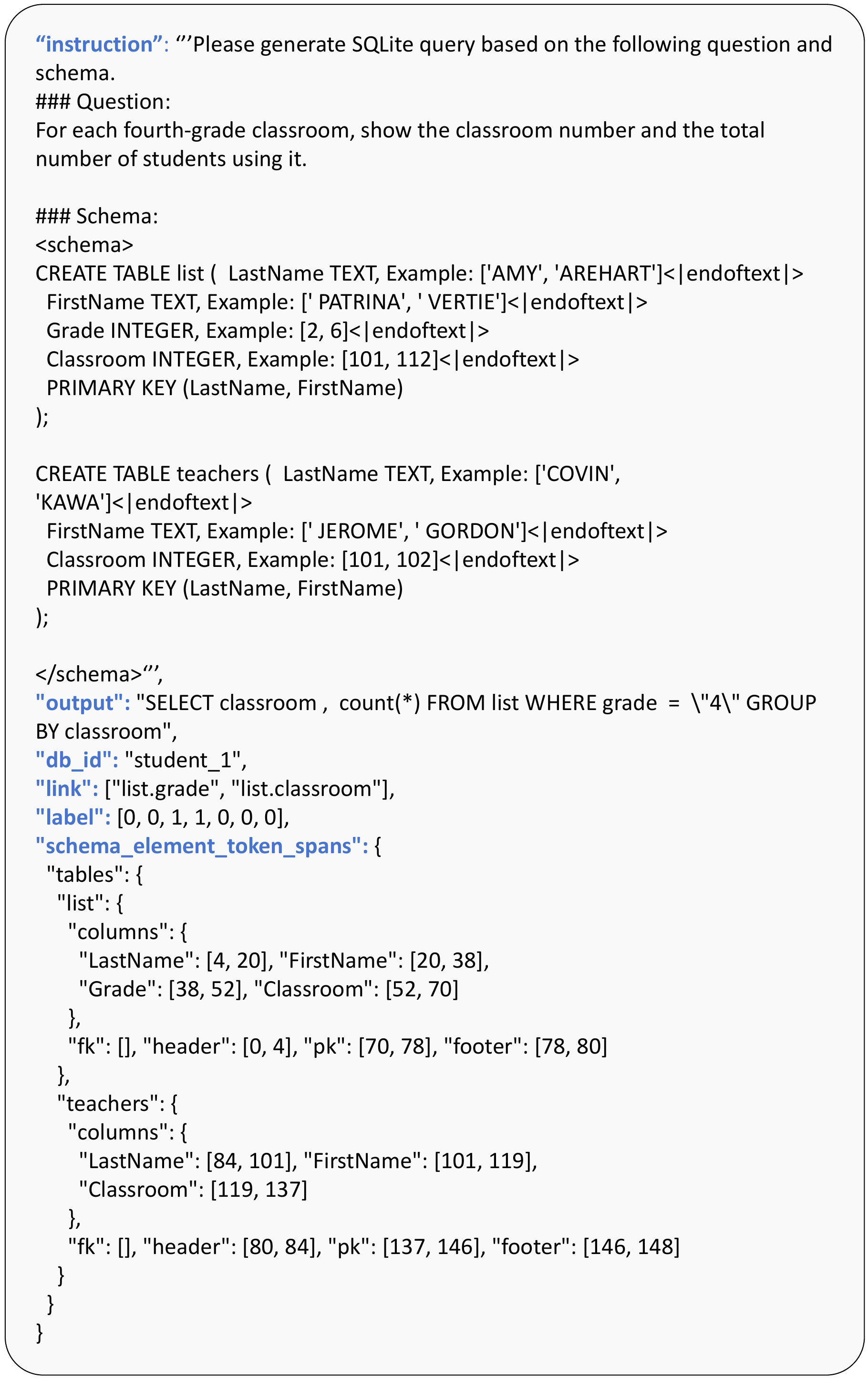}
\caption{An example of actual training data.
}
\label{data_example}
\end{figure}

Figure \ref{data_example} shows an example of the actual data we use. The \texttt{link} field contains the extracted ground-truth schema linking pairs. Based on this, we generate the final ground truth labels for schema linking (i.e., the \texttt{label} field).

We use \texttt{schema\_element\_token\_spans} field to record the positions of all schema elements. This is a nested dictionary structure that records the precise token spans for various elements within the schema. In addition to column definitions, it includes keys such as \texttt{fk} (foreign key definition span), \texttt{header} (table header definition span), \texttt{pk} (primary key definition span), and \texttt{footer} (table definition end span), which are used to accurately locate different components of the schema.

\subsection{Implementation Details}
\label{app:training_details}
\begin{table}[!htb]
\centering

\resizebox{1.0\columnwidth}{!}{
\begin{tabular}{lc}
\toprule[1pt]
\textbf{Hyperparameters} & \multicolumn{1}{l}{} \\ \hline
Epoch                    & 3(Spider),2(BIRD)    \\
Batchsize                & 1                    \\
Gradient Accumulation    & 6                    \\
Learning Rate            & 1.8e-5               \\
Weight Decay             & 1e-4                 \\
Max Grad Norm            & 1.0                  \\
Truncation Max Length    & 4096                 \\
LoRA Rank                & 64                   \\
LoRA Alpha               & 512                  \\
LoRA Dropout             & 0.08                 \\ \bottomrule[1pt]
\end{tabular}
}
\caption{Hyperparameters used for training.}
\label{Hyperparameters}

\end{table}

All our experiments were conducted on NVIDIA A30 GPUs with 24GB VRAM. The 7B models were trained on a single A30 GPU, while the 14B models used 2xA30 GPUs. The frameworks we used were PyTorch 2.5.1 and HuggingFace Transformers 4.51.3.

We employ LoRA \citep{lora} as the training method to reduce VRAM requirements. The target modules for LoRA are all linear layers in the LLM decoder. In addition to the LoRA trainable parameters, the linear layer weight \(W\) (Equation \ref{sigmoid}) is also part of the trainable parameters.

We consistently used the AdamW8bit optimizer \citep{8bit}, a cosine annealing learning rate scheduler, and enabled gradient checkpointing. We also utilized the Liger-Kernel \citep{liger-kernel} for further training efficiency optimization and enabled bfloat16 mixed-precision training. Other training hyperparameters are provided in Table \ref{Hyperparameters}. For SQL generation during inference, we employed greedy decoding.

\section{Case Study of Noisy Schema Sampling}
\label{sec:appendix_case_study}

\begin{figure}[!htb]
\centering

\includegraphics[width=0.5\textwidth]{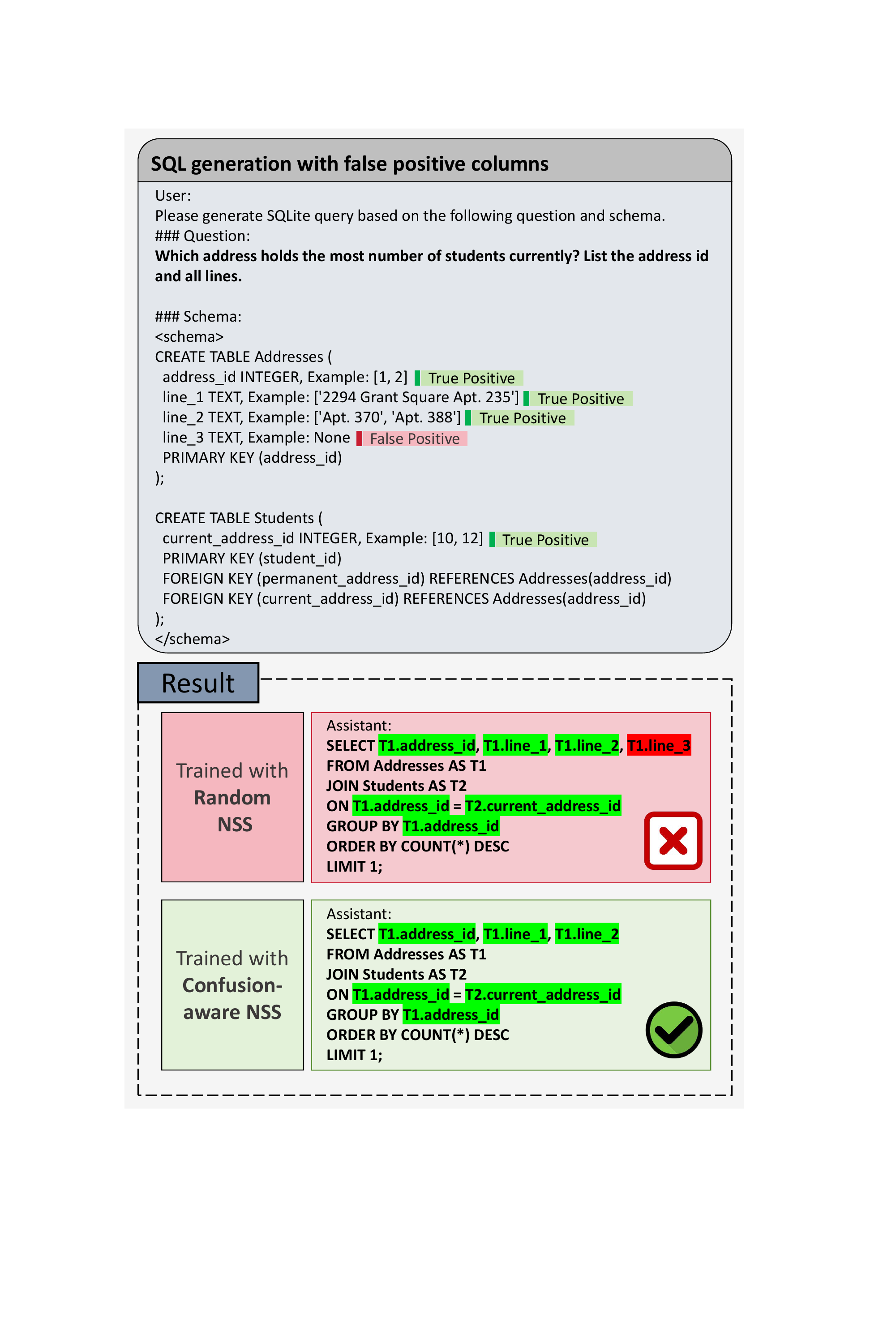}
\caption{A comparative case study of SQL generation with Random NSS versus Confusion-aware NSS when handling columns with False Positives. Green indicates True Positive columns, and red indicates False Positive columns. The model trained with Confusion-aware NSS (our method) correctly ignores the False Positive column \texttt{line\_3}, whereas Random NSS incorrectly includes it in the query.
}
\label{case_stu}
\end{figure}

To more intuitively demonstrate the robustness differences of various Noisy Schema Sampling (NSS) strategies in handling False Positive (FP) columns introduced by schema linking in SQL generation tasks, we conduct a case study, as illustrated in Figure \ref{case_stu}. In this case, the user's question is: \emph{"Which address holds the most number of students currently? List the address id and all lines."}

The provided database schema includes an \emph{Addresses} table and a \texttt{Students} table. In the \texttt{Addresses} table, \texttt{address\_id}, \texttt{line\_1}, and \texttt{line\_2} are True Positive (TP) columns relevant to the question, while \texttt{line\_3} is a False Positive (FP) column. Its example value is \texttt{None}, indicating it is an empty column with no data, which was mistakenly identified as relevant during the schema linking.

We compare the model performance under two NSS training strategies:
(1) \textbf{Model trained with Random NSS}: The generated SQL query incorrectly includes the FP column \texttt{line\_3}. This indicates that despite being exposed to random noise during training, the model failed to adequately learn to ignore such misleading FP columns, especially when the column \texttt{line\_3} has some textual similarity to "\emph{all lines}" in the question.
(2) \textbf{Model trained with Confusion-aware NSS}: The generated SQL query correctly ignores the FP column \texttt{line\_3} and selects only the truly relevant columns. This demonstrates that Confusion-aware NSS, by enabling the model to focus on learning its own 'easily confused' noisy patterns during training, can more effectively enhance the model's robustness to schema linking errors.

\end{document}